\documentclass[conference]{IEEEtran}
\IEEEoverridecommandlockouts
\usepackage{cite}
\usepackage{amsmath,amssymb,amsfonts,bm}
\usepackage{algorithmic}
\usepackage{graphicx}
\usepackage{textcomp}
\usepackage{xcolor}
\usepackage{balance}
\usepackage{pifont}
\usepackage{wasysym}
\usepackage{algorithm,algorithmic}
\def\BibTeX{{\rm B\kern-.05em{\sc i\kern-.025em b}\kern-.08em
    T\kern-.1667em\lower.7ex\hbox{E}\kern-.125emX}}
\begin{document}

\title{Motion Semantics Guided Normalizing Flow for Privacy-Preserving Video Anomaly Detection 
}

\author{
    Yang Liu$^{1}$, 
    Boan Chen$^{2}$, 
    Yuanyuan Meng$^{3}$, 
    Jing Liu$^{4}$, 
    Zhengliang Guo$^{5}$, 
    Wei Zhou$^{6}$, 
    Peng Sun$^{7,*}$, 
    Hong Chen$^{1}$ \\ 

    $^{1}$Tongji Univ. \quad $^{2}$SJTU \quad $^{3}$Shanghai Creative Studies Institute \quad $^{4}$UBC \quad $^{5}$Fudan Univ. \quad $^{6}$Cardiff Univ. \quad $^{7}$DKU \\

    yang\_liu@ieee.org, bchen4@sjtu.edu.cn, peng.sun568@duke.edu, chenhong2019@tongji.edu.cn
}

\maketitle
\begin{abstract}
As embodied perception systems increasingly bridge digital and physical realms in interactive multimedia applications, the need for privacy-preserving approaches to understand human activities in physical environments has become paramount. Video anomaly detection is a critical task in such embodied multimedia systems for intelligent surveillance and forensic analysis. Skeleton-based approaches have emerged as a privacy-preserving alternative that processes physical world information through abstract human pose representations while discarding sensitive visual attributes such as identity and facial features. However, existing skeleton-based methods predominantly model continuous motion trajectories in a monolithic manner, failing to capture the hierarchical nature of human activities composed of discrete semantic primitives and fine-grained kinematic details, which leads to reduced discriminability when anomalies manifest at different abstraction levels. In this regard, we propose Motion Semantics Guided Normalizing Flow (MSG-Flow) that decomposes skeleton-based VAD into hierarchical motion semantics modeling. It employs vector quantized variational auto-encoder to discretize continuous motion into interpretable primitives, an autoregressive Transformer to model semantic-level temporal dependencies, and a conditional normalizing flow to capture detail-level pose variations. Extensive experiments on benchmarks (HR-ShanghaiTech \& HR-UBnormal) demonstrate that MSG-Flow achieves state-of-the-art performance with 88.1\% and 75.8\% AUC respectively.
\end{abstract}


\section{Introduction}
\label{sec:intro}

The convergence of embodied intelligence and interactive multimedia systems is fundamentally transforming how we process and understand physical world information \cite{xie2025acorn,su2025large}. Embodied perception, which emphasizes the acquisition and understanding of information through active interaction with real environments, has become crucial for modern multimedia applications including intelligent surveillance, human-robot collaboration, and smart home systems~\cite{liu2025networking,huang2023multi,liu2026edge}. Video Anomaly Detection (VAD) represents a quintessential embodied perception task, requiring systems to continuously perceive, process, and interpret human activities in dynamic physical spaces \cite{liu2022collaborative,guo2025adaptive,cheng2024normality,liu2024generalized}. As cities worldwide deploy extensive camera networks generating petabytes of visual data daily, the demand for intelligent systems capable of real-time anomaly screening in these unstructured physical settings has intensified~\cite{cheng2024denoising}. However, traditional appearance-based VAD methods that operate on raw RGB frames raise significant privacy concerns, as they process and potentially store sensitive biometric information including facial features, clothing patterns, and body characteristics~\cite{liu2022learning,liu2025improving}. The conflict between security needs and privacy rights in embodied multimedia systems has motivated the development of privacy-preserving approaches \cite{liu2025m2s2l,yang2025medaide,qian2026survey} that extract task-relevant kinematic information while discarding identifiable attributes. Skeleton-based VAD, which represents humans as abstract skeletal graphs of joint coordinates, offers an elegant solution by focusing solely on motion kinematics and maintaining physical grounding through pose-based representations~\cite{liu2023amp,liu2024diffskill,zhao2026new}.

Existing skeleton-based VAD methods can be categorized into three paradigms: reconstruction-based, prediction-based, and density estimation approaches~\cite{liu2023learning,luo2021normal,hirschorn2023normalizing}. Reconstruction-based methods~\cite{morais2019learning,markovitz2020graph} train autoencoders on normal samples and detect anomalies via reconstruction errors, assuming abnormal patterns are difficult to reconstruct. Prediction-based methods~\cite{li2024stnmamba,rodrigues2020multi} forecast future poses from historical observations, flagging deviations between predictions and actual trajectories as anomalies. Recent advances leverage normalizing flows~\cite{hirschorn2023normalizing,wu2024daflow} to directly model probability density of skeleton sequences, offering theoretically grounded anomaly scoring through exact likelihood computation. Despite these efforts, a critical limitation persists: existing methods treat skeleton sequences as continuous trajectories in high-dimensional space, modeling them monolithically without considering the inherent hierarchical structure of human motion. Human activities in physical environments naturally decompose into discrete semantic units (e.g., basic actions like ``raising arm,'' ``stepping forward'') that compose into complex behaviors through temporal sequencing~\cite{kuehne2014language}. Anomalies in embodied contexts may manifest at different abstraction levels, either as illogical combinations of otherwise normal actions (i.e., semantic anomalies) or as distorted execution of recognizable actions (i.e., kinematic anomalies)~\cite{liu2025crcl}. Current approaches lack mechanisms to disentangle and separately model these two facets, limiting both detection performance and interpretability in interactive multimedia systems.

Motivated by the compositional nature of human motions in physical environments \cite{liu2022appearance,xie2025haer}, we propose Motion Semantics Guided Normalizing Flow (MSG-Flow), a hierarchical framework that explicitly addresses skeleton-based VAD at two interconnected levels suitable for embodied perception systems. At the semantic level, we discretize continuous motion into a finite vocabulary of learned primitives using Vector Quantized Variational Auto-Encoder (VQ-VAE)~\cite{vqvae}, transforming each skeleton sequence into a symbolic representation analogous to words forming sentences in natural language. Our discrete representation enables efficient processing in resource-constrained embodied systems. An autoregressive Transformer then captures temporal dependencies among these primitives, learning which action sequences constitute normal behavior patterns in physical spaces. At the kinematic level, a conditional normalizing flow models the distribution of residual details (i.e., subtle pose variations that distinguish individual execution styles) conditioned on the inferred primitive sequence, enabling robust perception under the inherent uncertainty and variability of real-world physical interactions.

The main contributions of this work are as follows:
\begin{itemize}
    \item We introduce a hierarchical framework for skeleton-based VAD that explicitly models motion at discrete semantic and continuous kinematic levels, addressing the needs of embodied perception in interactive multimedia systems.
    \item We develop MSG-Flow that integrates vector quantization, autoregressive transformers, and conditional normalizing flows to capture multi-level human motion patterns while maintaining computational efficiency suitable for embodied applications.
    \item Extensive experiments on HR-ShanghaiTech and HR-UBnormal benchmarks demonstrate that MSG-Flow achieves state-of-the-art performance while maintaining model compactness comparable to existing methods, validating its effectiveness in open and physical spaces.
\end{itemize}

\section{Methodology}
\label{sec:method}

\subsection{Overview}

While previous approaches~\cite{liu2025privacy} treat skeletal sequences as continuous trajectories in high-dimensional space, human motion in physical environments inherently exhibits a compositional structure where behaviors emerge from the combination of basic movement patterns, or \emph{primitives}. MSG-Flow decomposes the VAD task into two interconnected modeling levels: a discrete primitive level that captures abstract motion semantics, and a continuous detail level that models fine-grained pose variations. 

\begin{figure}[t]
\centering
\includegraphics[width=\columnwidth]{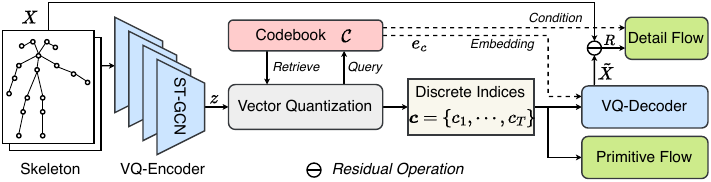}
\caption{Overview of MSG-Flow framework. The input skeleton sequence is encoded into discrete motion primitives through VQ-VAE. The primitive sequence is modeled by an autoregressive Transformer (Primitive Flow), while residual details are modeled by a conditional normalizing flow (Detail Flow).}
\label{fig:overview}
\end{figure}

Fig.~\ref{fig:overview} illustrates the overall architecture of MSG-Flow. Given an input skeleton sequence $\mathbf{X} = \{\mathbf{x}_t\}_{t=1}^T$ where $\mathbf{x}_t \in \mathbb{R}^{V \times 2}$ represents the 2D coordinates of $V$ joints at time $t$, the framework operates through three cascaded components. First, the VQ-VAE encodes the continuous motion into a discrete sequence of primitive indices $\mathbf{c} = \{c_t\}_{t=1}^T$ with $c_t \in \{1, 2, \ldots, K\}$, which compresses the input into a compact semantic representation where each index corresponds to a learnable motion primitive stored in codebook $\mathcal{C} = \{\mathbf{e}_k\}_{k=1}^K$. The discretization enables efficient symbolic processing and interpretability. Second, the \emph{Primitive Flow}, an autoregressive Transformer-based model, captures temporal dependencies among motion primitives and estimates the probability distribution $p(\mathbf{c})$ over primitive sequences. Third, the \emph{Detail Flow}, implemented as a conditional normalizing flow, models the residual differences $\mathbf{R} = \mathbf{X} - \tilde{\mathbf{X}}$ between the original sequence and its primitive-level reconstruction $\tilde{\mathbf{X}}$, capturing subtle kinematic variations that distinguish individual execution styles.

\subsection{Motion Primitive Learning via Vector Quantization}

The foundation of our hierarchical modeling lies in discovering and representing motion primitives from continuous skeletal trajectories. We employ a VQ-VAE~\cite{vqvae} that learns to discretize the continuous motion space into a finite set of representative primitives. 

The VQ-encoder maps input sequence $\mathbf{X} \in \mathbb{R}^{T \times V \times 2}$ to a sequence of continuous latent vectors through the Spatio-Temporal Graph Convolutional Network (ST-GCN):
\begin{equation}
\mathbf{Z} = \text{Enc}(\mathbf{X}; \theta_e) \in \mathbb{R}^{T \times d}
\end{equation}
where $\theta_e$ denotes encoder parameters and $d$ is the latent dimension. The encoder consists of $L=4$ stacked ST-GCN layers~\cite{stgcn}, each comprising spatial graph convolution followed by temporal convolution. To preserve temporal ordering, we augment inputs with sinusoidal positional encodings~\cite{transformer}.

Each latent vector $\mathbf{z}_t \in \mathbb{R}^d$ is quantized to its nearest neighbor in codebook $\mathcal{C}$ using Euclidean distance:
\begin{equation}
c_t = \arg\min_{k \in \{1,\ldots,K\}} \|\mathbf{z}_t - \mathbf{e}_k\|_2
\end{equation}

The quantized representation $\mathbf{z}_q = \mathbf{e}_{c_t}$ replaces the continuous encoding during forward pass. To enable gradient-based learning despite non-differentiable quantization, we employ the straight-through estimator~\cite{vqvae} that copies gradients:
\begin{equation}
\mathbf{z}_q = \mathbf{z} + \text{sg}[\mathbf{e}_{c_t} - \mathbf{z}]
\end{equation}
where $\text{sg}[\cdot]$ denotes the stop-gradient operator.

The codebook is learned through Exponential Moving Average (EMA) updates rather than direct gradient descent~\cite{vqvae}. For each entry $\mathbf{e}_k$, we maintain running statistics:
\begin{equation}
\begin{split}
N_k^{(t)} &= \gamma N_k^{(t-1)} + (1-\gamma) \sum_{i=1}^{B} \mathbb{1}[c_i = k] \\
\mathbf{m}_k^{(t)} &= \gamma \mathbf{m}_k^{(t-1)} + (1-\gamma) \sum_{i=1}^{B} \mathbb{1}[c_i = k] \mathbf{z}_i
\end{split}
\end{equation}
where $\mathbb{1}[\cdot]$ is the indicator function, $B$ is batch size and $\gamma = 0.99$. The codebook vectors are updated as $\mathbf{e}_k = \mathbf{m}_k / N_k$.

The decoder reconstructs coarse skeleton sequence from quantized primitives: $\tilde{\mathbf{X}} = \text{Dec}(\{\mathbf{e}_{c_1}, \ldots, \mathbf{e}_{c_T}\}; \theta_d)$. It employs a Transformer architecture that attends to the entire primitive sequence, generating smooth reconstructions capturing global motion patterns. Each time step's output is obtained via: $\tilde{\mathbf{x}}_t = \text{MLP}(\mathbf{h}_t^{\text{dec}}) \in \mathbb{R}^{V \times 2}$.

\subsection{Motion Semantics Modeling with Primitive Flow}

Once continuous sequences are discretized into primitive indices, we model their probability distribution to capture the temporal structure inherent in normal activities. In physical environments, certain primitives naturally follow others (e.g., ``bend knees'' before ``jump''), while some transitions are rare or physically implausible (e.g., ``lying down'' immediately before ``running''). The Primitive Flow captures these dependencies through autoregressive modeling:
\begin{equation}
p(\mathbf{c}) = \prod_{t=1}^{T} p(c_t | c_{<t}; \theta_p)
\label{eq:primitive_flow}
\end{equation}
where $c_{<t} = \{c_1, \ldots, c_{t-1}\}$ denotes preceding primitives. The proposed factorization enables efficient computation while naturally encoding temporal causality. 

The conditional distributions are parameterized by a Transformer decoder~\cite{gpt}. Specifically, input primitive indices are first embedded and combined with positional encodings: $\mathbf{E}_t = \text{Embed}(c_t) + \text{PE}(t)$, where $\text{Embed:} \{1,\ldots,K\} \rightarrow \mathbb{R}^{d_{\text{model}}}$ is learnable. The embeddings are then processed through $N_{\text{layer}}$ decoder layers, each with masked self-attention and feed-forward sublayers. The masked attention ensures position $t$ can only attend to positions $\leq t$ via causal mask $\mathbf{M} \in \mathbb{R}^{T \times T}$ where $\mathbf{M}_{ij} = 0$ if $i \geq j$, else $-\infty$. Output hidden states are projected to probability distribution over $K$ primitives:
\begin{equation}
p(c_t | c_{<t}) = \text{softmax}(\mathbf{W}_{\text{out}} \mathbf{h}_t^{(N_{\text{layer}})} + \mathbf{b}_{\text{out}})
\end{equation}

\subsection{Fine-Grained Variations Modeling with Detail Flow}

While Primitive Flow captures high-level semantics, many anomalies in physical spaces manifest as subtle pose deviations not reflected in primitive sequences. For instance, a person may execute a ``walking'' primitive with unusual joint angles or gait patterns. The Detail Flow, a conditional normalizing flow~\cite{glow}, models the distribution of residuals $\mathbf{R} = \mathbf{X} - \tilde{\mathbf{X}}$ conditioned on primitive sequence $\mathbf{c}$, capturing these fine-grained kinematic variations.

A normalizing flow defines bijective mapping $f: \mathcal{X} \rightarrow \mathcal{Z}$ transforming data from complex to simple base distribution $p_Z(\mathbf{z}) = \mathcal{N}(\mathbf{z}; \mathbf{0}, \mathbf{I})$. The probability density follows change-of-variables:
\begin{equation}
\log p(\mathbf{r}) = \log p_Z(f(\mathbf{r})) + \log \left|\det \frac{\partial f}{\partial \mathbf{r}}\right|
\label{eq:flow}
\end{equation}

To ensure tractability, transformation $f$ comprises $K_{\text{flow}}=8$ invertible coupling layers: $f = f^{(K_{\text{flow}})} \circ \cdots \circ f^{(1)}$. Each layer splits input into halves and applies affine transformation to one half based on the other:
\begin{equation}
\begin{split}
\mathbf{r}_0, \mathbf{r}_1 &= \text{split}(\mathbf{r}) \\
\mathbf{s}, \mathbf{t} &= \text{NN}(\mathbf{r}_0; \mathbf{e}_c) \\
\mathbf{r}_1' &= \mathbf{r}_1 \odot \exp(\mathbf{s}) + \mathbf{t}
\end{split}
\end{equation}
where scale and translation parameters $\mathbf{s}, \mathbf{t}$ are predicted by a Neural Network ($\text{NN}$), and $\odot$ denotes element-wise multiplication.

Conditioning on primitive embeddings $\mathbf{e}_c$ is implemented through Conditional Batch Normalization (CBN)~\cite{cbn} within $\text{NN}$. In CBN, affine parameters are predicted from conditioning: $\gamma(\mathbf{e}_c) = \mathbf{W}_{\gamma} \mathbf{e}_c + \mathbf{b}_{\gamma}$ and $\beta(\mathbf{e}_c) = \mathbf{W}_{\beta} \mathbf{e}_c + \mathbf{b}_{\beta}$, yielding:
\begin{equation}
\text{CBN}(\mathbf{h}; \mathbf{e}_c) = \gamma(\mathbf{e}_c) \cdot \frac{\mathbf{h} - \mu}{\sqrt{\sigma^2 + \epsilon}} + \beta(\mathbf{e}_c)
\end{equation}
It allows flow transformation to adapt parameters based on processed primitive, enabling modeling of different residual distributions for different motion contexts.

\subsection{Training Objectives and Strategy}

The training of MSG-Flow jointly optimizes VQ-VAE, Primitive Flow, and Detail Flow, and the total loss is:
\begin{equation}
\mathcal{L}_{\text{total}} = \mathcal{L}_{\text{recon}} + \lambda \mathcal{L}_{\text{commit}} + \mathcal{L}_{\text{prim}} + \mathcal{L}_{\text{detail}}
\label{eq:total_loss}
\end{equation}
The reconstruction loss ensures VQ-VAE encodes and decodes with minimal information loss: $\mathcal{L}_{\text{recon}} = \mathbb{E}_{\mathbf{X}}[\|\mathbf{X} - \text{Dec}(\mathbf{e}_{c_1}, \ldots, \mathbf{e}_{c_T})\|_2^2]$. The commitment loss $\mathcal{L}_{\text{commit}} = \mathbb{E}[\|\text{sg}[\mathbf{Z}] - \mathbf{E}_c\|_2^2]$ encourages encoder outputs to stay close to assigned codebook entries, with $\lambda = 0.25$. The primitive flow loss maximizes log-likelihood of observed sequences: $\mathcal{L}_{\text{prim}} = -\mathbb{E}_{\mathbf{c}}[\sum_{t=1}^{T} \log p(c_t | c_{<t})]$. The detail flow loss trains conditional normalizing flow: $\mathcal{L}_{\text{detail}} = -\mathbb{E}_{\mathbf{X}, \mathbf{c}}[\log p(\mathbf{R} | \mathbf{c})]$, where $\log p(\mathbf{R} | \mathbf{c}) = \log p_Z(f(\mathbf{R}; \mathbf{c})) + \sum_{k=1}^{K_{\text{flow}}} \log|\det \frac{\partial f^{(k)}}{\partial f^{(k-1)}}|$.

\subsection{Anomaly Detection}

Given test sequence $\mathbf{X}^{\text{test}}$, MSG-Flow computes anomaly score quantifying deviation from learned normal motion distribution. We first infer the most likely primitive sequence by encoding through VQ-encoder: $\mathbf{Z}^{\text{test}} = \text{Enc}(\mathbf{X}^{\text{test}})$ and quantizing to nearest codebook entries: $c_t^* = \arg\min_{k} \|\mathbf{z}_t^{\text{test}} - \mathbf{e}_k\|_2$ for $t = 1, \ldots, T$. The semantic-level anomaly score is computed as negative log-probability under Primitive Flow:
\begin{equation}
s_{\text{prim}} = -\log p(\mathbf{c}^*) = -\sum_{t=1}^{T} \log p(c_t^* | c_{<t}^*)
\label{eq:semantic_score}
\end{equation}
For detail-level scoring, we reconstruct prototype trajectory $\tilde{\mathbf{X}}^{\text{test}} = \text{Dec}(\mathbf{e}_{c_1^*}, \ldots, \mathbf{e}_{c_T^*})$ and compute residual $\mathbf{R}^{\text{test}} = \mathbf{X}^{\text{test}} - \tilde{\mathbf{X}}^{\text{test}}$. The detail score evaluates residual probability under conditional Detail Flow:
\begin{equation}
\begin{split}
s_{\text{detail}} = &-\log p(\mathbf{R}^{\text{test}} | \mathbf{c}^*) \\
= &-\log p_Z(\mathbf{Z}^{\text{test}}) - \sum_{k=1}^{K_{\text{flow}}} \log|\det J_{f^{(k)}}|
\end{split}
\label{eq:detail_score}
\end{equation}
where $\mathbf{Z}^{\text{test}} = f(\mathbf{R}^{\text{test}}; \mathbf{c}^*)$. 
The total anomaly score combines both components: $s_{\text{total}} = \alpha \cdot s_{\text{prim}} + (1-\alpha) \cdot s_{\text{detail}}$. We employ adaptive mechanism based on primitive inference confidence: $\text{conf} = \frac{1}{T}\sum_{t=1}^{T} \max_{k} p(c_t = k | \mathbf{z}_t^{\text{test}})$
yielding adaptive weight $\alpha = \sigma(5.0 \cdot (\text{conf} - 0.5))$ where $\sigma(\cdot)$ is sigmoid function. When primitive inference is confident, the model relies more on semantic scoring; otherwise, more on detail scoring. Our adaptive fusion enables robust anomaly detection under varying levels of motion ambiguity commonly encountered in real-world physical environments.

\section{Experiments}
\label{sec:experiments}

\subsection{Experimental Setup}

\subsubsection{Datasets}
We evaluate MSG-Flow on two skeleton-based VAD benchmarks following the human-related (HR) protocol~\cite{flaborea2024contracting}: 

\textbf{HR-ShanghaiTech}~\cite{liu2018future,flaborea2024contracting} is derived from the ShanghaiTech Campus dataset by retaining only anomalies caused by human actions while excluding object-related events. The HR subset contains 297,000 normal frames and 16,122 abnormal frames in the test set. 

\textbf{HR-UBnormal}~\cite{acsintoae2022ubnormal,flaborea2024contracting} is a synthetic dataset created with Cinema4D, including 147,887 normal frames and 86,864 abnormal frames across diverse synthetic scenes. The synthetic nature provides controlled evaluation conditions while maintaining diversity in anomaly types. 

\subsubsection{Evaluation Metrics}

The primary metric is the Area Under the Receiver Operating Characteristic Curve (AUC) computed at frame granularity, measuring the ability to distinguish normal from abnormal frames across all possible decision thresholds. AUC provides a threshold-independent assessment of detection performance, which is crucial for practical deployment in embodied systems where optimal operating points may vary by application context.
In addition, we report the \textbf{number of trainable parameters} (in millions) and \textbf{floating-point operations} (FLOPs, in giga-operations) required for processing a single 24-frame skeleton sequence.

\subsubsection{Implementation Details}

MSG-Flow is implemented in PyTorch 2.0 and trained on a single NVIDIA RTX 3090 GPU. The VQ-VAE uses a codebook with $K=512$ entries and latent dimension $d=256$. The ST-GCN encoder has 4 layers with channel dimensions $\{64, 64, 128, 256\}$. The Primitive Flow Transformer comprises 6 layers with $d_{\text{model}}=256$ and 8 attention heads. The Detail Flow consists of 8 coupling layers operating on intermediate dimension 128. We process skeleton sequences using sliding windows of $24$ frames with stride 12, providing sufficient temporal context to capture motion primitives while enabling efficient processing.

\subsection{Comparison with State-of-the-Art Methods}

Table~\ref{tab:comparison} compares MSG-Flow against recent skeleton-based VAD methods. Our approach achieves superior performance on both benchmarks, attaining 88.1\% AUC on HR-ShanghaiTech and 75.8\% AUC on HR-UBnormal. Compared to the previous best-performing method (DA-Flow~\cite{wu2024daflow}), MSG-Flow improves by +0.3\% and +1.6\% respectively, demonstrating the effectiveness of hierarchical motion semantics modeling. 

The substantial gain on HR-UBnormal is particularly noteworthy, as this dataset contains diverse anomaly patterns in open-set evaluation scenarios that better reflect real-world physical environments. It validates MSG-Flow's capability to capture both semantic-level anomalies (unusual action sequences) and kinematic-level anomalies (distorted pose executions), which is essential for robust anomaly detection in unstructured physical spaces characteristic of embodied perception applications.

MSG-Flow achieves good results while maintaining reasonable model complexity with 10.3M parameters and 1.2G FLOPs. The parameter increase relative to lightweight methods~\cite{wu2024daflow} is justified by the hierarchical architecture with enhanced representation capacity, as evidenced by superior detection performance across different evaluation contexts.

\begin{table}[t]
\centering
\caption{Comparison with existing skeleton-based VAD methods.}
\label{tab:comparison}
\resizebox{\columnwidth}{!}{
\begin{tabular}{lccc}
\hline
\textbf{Method} & \textbf{Venue} & \textbf{HR-STC (\%)} & \textbf{HR-UBnormal (\%)} \\
\hline
MPED-RNN~\cite{morais2019learning} & CVPR'19 & 75.4 & 61.2 \\
GEPC~\cite{markovitz2020graph} & CVPR'20 & 74.8 & 55.2 \\
Multi-Time~\cite{rodrigues2020multi} & WACV'20 & 77.0 & -- \\
MoCoDAD~\cite{flaborea2023mocodad} & ICCV'23 & 77.6 & 68.4 \\
STG-NF~\cite{hirschorn2023normalizing} & ICCV'23 & 87.4 & 71.8 \\
COSKAD-Eucl~\cite{flaborea2024contracting} & PR'24 & 77.1 & 65.0 \\
COSKAD-Hyp~\cite{flaborea2024contracting} & PR'24 & 75.6 & 65.5 \\
TrajREC~\cite{stergiou2024trajrec} & WACV'24 & 77.9 & 68.2 \\
DA-Flow~\cite{wu2024daflow} & TMM'25 & \underline{87.8} & \underline{74.2} \\
\hline
\textbf{MSG-Flow (Ours)} & -- & \textbf{88.1} & \textbf{75.8} \\
\hline
\end{tabular}}
\end{table}

\subsection{Ablation Studies}

\subsubsection{Component Ablation}

To validate the contribution of each component in MSG-Flow's hierarchical architecture, we conduct systematic ablation studies presented in Table~\ref{tab:ablation_component}. We evaluate five variants: (1) \textbf{VQ-VAE Only}: Using only reconstruction error from VQ-VAE decoder without flow-based modeling, serving as a baseline for vector quantization effectiveness; (2) \textbf{w/o Primitive Flow}: Removing the semantic-level modeling, scoring only via Detail Flow on raw residuals; (3) \textbf{w/o Detail Flow}: Removing kinematic-level modeling, scoring only via Primitive Flow probabilities; (4) \textbf{Fixed $\alpha$}: Replacing adaptive fusion weight with fixed $\alpha=0.5$; (5) \textbf{Full Model}: Complete MSG-Flow with all components.

First, the VQ-VAE-only baseline (82.3\% / 68.5\%) demonstrates that discretization alone provides reasonable anomaly detection through reconstruction-based scoring, but lacks the discriminative power of density-based modeling. It validates that motion primitive learning captures meaningful semantic structure, though not sufficient for state-of-the-art performance.

Second, removing either flow component causes significant performance drops, confirming that both semantic and kinematic modeling are necessary for comprehensive anomaly detection. Removing Primitive Flow results in 85.2\% / 72.1\% performance, while removing Detail Flow yields 83.7\% / 69.8\%. The larger degradation on HR-UBnormal when removing Primitive Flow (-3.7\%) suggests that semantic-level anomalies (unusual action sequences) are more prevalent in this diverse synthetic dataset, highlighting the importance of temporal dependency modeling for open-set scenarios.

Third, the adaptive fusion mechanism (vs. fixed $\alpha$) contributes +0.7\% and +0.9\% gains by dynamically weighting semantic and kinematic scores based on primitive inference confidence. The adaptive strategy is particularly beneficial in physical environments where motion ambiguity varies across different activities and contexts.

\begin{table}[t]
    \centering
    \caption{Results of component ablation study.}
    \label{tab:ablation_component}
    \resizebox{\columnwidth}{!}{
    \begin{tabular}{lccc}
    \hline
    \textbf{Variant} & \textbf{HR-STC (\%)} & \textbf{HR-UBnormal (\%)} & \textbf{Params (M)} \\
    \hline
    VQ-VAE Only & 82.3 & 68.5 & 4.9 \\
    w/o Primitive Flow & 85.2 & 72.1 & 5.3 \\
    w/o Detail Flow & 83.7 & 69.8 & 5.3 \\
    Fixed $\alpha$ & 87.4 & 74.9 & 10.3 \\
    \hline
    \textbf{Full Model} & \textbf{88.1} & \textbf{75.8} & 10.3 \\
    \hline
    \end{tabular}}
\end{table}

\subsubsection{Loss Function Ablation}

Table~\ref{tab:ablation_loss} examines the impact of each loss in Eq.~\eqref{eq:total_loss}. We evaluate four variants with different loss configurations: (1) \textbf{w/o $\mathcal{L}_{\text{commit}}$}: Removing commitment loss to assess its role in codebook stability; (2) \textbf{w/o $\mathcal{L}_{\text{recon}}$}: Removing reconstruction loss to examine its necessity for learning meaningful primitives; (3) \textbf{$\lambda=0.1$} and \textbf{$\lambda=0.5$}: Testing different commitment loss weights to understand the balance between encoder flexibility and codebook utilization; (4) \textbf{Full Loss}: Using all components with $\lambda=0.25$.

The commitment loss is crucial for maintaining codebook diversity and preventing code collapse. Without it, only 62.3\% of codebook entries are actively used during training, indicating severe underutilization where many primitives remain unlearned. It leads to reduced representation capacity and performance degradation (86.5\% / 73.8\%). 
Similarly, removing reconstruction loss impairs VQ-VAE's ability to learn meaningful primitives, resulting in 85.9\% / 72.4\% performance despite high codebook usage (89.7\%). It suggests that without reconstruction guidance, the learned primitives may not effectively capture motion semantics.
The commitment weight $\lambda$ requires careful tuning to balance encoder flexibility with codebook utilization.

\begin{table}[t]
    \centering
    \caption{Results of loss function ablation study.}
    \label{tab:ablation_loss}
    \resizebox{\columnwidth}{!}{
    \begin{tabular}{lccc}
    \hline
    \textbf{Config.} & \textbf{HR-STC (\%)} & \textbf{HR-UBnormal (\%)} & \textbf{Codebook Usage (\%)} \\
    \hline
    w/o $\mathcal{L}_{\text{commit}}$ & 86.5 & 73.8 & 62.3 \\
    w/o $\mathcal{L}_{\text{recon}}$ & 85.9 & 72.4 & 89.7 \\
    $\lambda=0.1$ & 87.6 & 75.2 & 78.4 \\
    $\lambda=0.5$ & 87.3 & 74.7 & 91.2 \\
    \hline
    \textbf{Full ($\lambda=0.25$)} & \textbf{88.1} & \textbf{75.8} & \textbf{85.5} \\
    \hline
    \end{tabular}}
\end{table}

\subsubsection{Parameter Sensitivity Analysis}

\begin{figure}[t]
\centering
\includegraphics[width=\columnwidth]{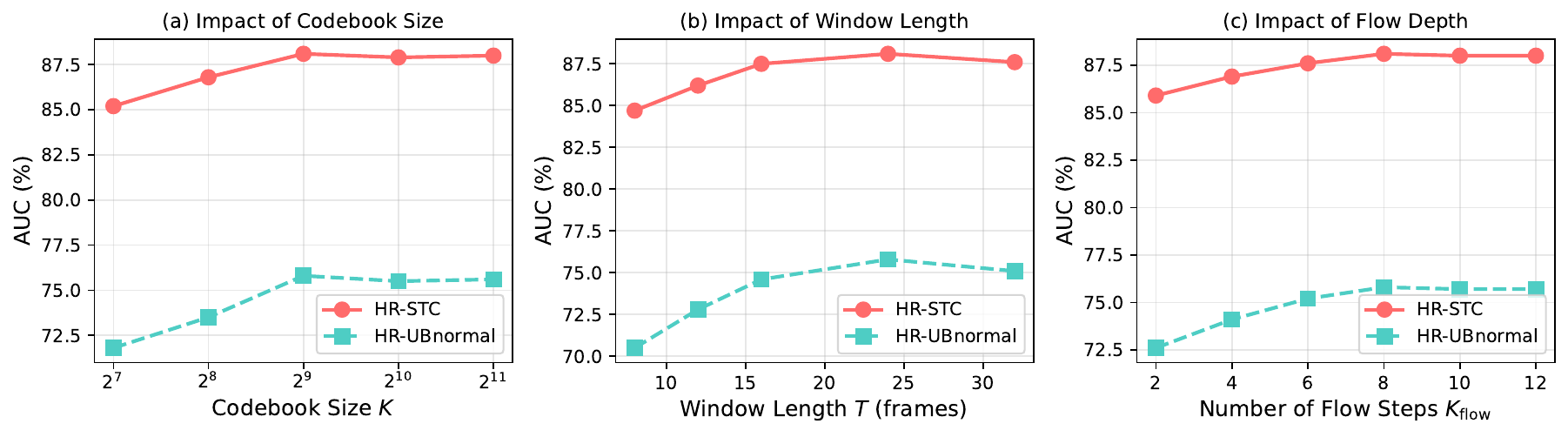}
\caption{Sensitivity analysis on (a) Codebook size $K$, (b) Window length $T$, and (c) Flow steps $K_{\text{flow}}$. Solid/dashed lines: HR-STC/HR-UBnormal.}
\label{fig:sensitivity}
\end{figure}

We analyze the sensitivity of MSG-Flow to three key hyperparameters: codebook size $K$, window length $T$, and number of flow steps $K_{\text{flow}}$, as illustrated in Fig.~\ref{fig:sensitivity}(a)-(c), respectively.

\textbf{Codebook Size.} We evaluate $K \in \{128, 256, 512, 1024, 2048\}$. Performance improves from $K=128$ (84.2\% / 71.3\%) to $K=512$ (88.1\% / 75.8\%) as larger codebooks provide finer-grained motion primitives capable of capturing more diverse action patterns. However, further increasing to $K=2048$ (87.9\% / 75.3\%) yields diminishing returns and risks overfitting, as the model struggles to populate excessively large codebooks with limited training data. It suggests that $K=512$ provides an optimal balance between representation capacity and generalization, capturing sufficient diversity in motion primitives without overfitting.

\textbf{Window Length.} We test $T \in \{8, 12, 16, 20, 24, 28, 32\}$ frames. Shorter windows ($T=8$: 84.7\% / 70.5\%) lack sufficient temporal context to capture complete motion primitives, as many basic actions span longer durations. Performance peaks at $T=24$ frames (88.1\% / 75.8\%), which corresponds to approximately 1 second at typical frame rates, providing adequate context to recognize most atomic action units while maintaining computational efficiency. Longer windows ($T=32$: 87.6\% / 75.1\%) increase computational cost without performance benefits, as they exceed typical primitive durations and may introduce noise from multiple overlapping primitives.

\textbf{Flow Depth.} We vary $K_{\text{flow}} \in \{2, 4, 6, 8, 10, 12\}$. Shallow flows ($K_{\text{flow}}=2$: 85.9\% / 73.2\%) have insufficient expressiveness to model complex residual distributions. Performance improves with depth, saturating around $K_{\text{flow}}=8$ (88.1\% / 75.8\%). Deeper flows ($K_{\text{flow}}=12$: 88.0\% / 75.7\%) offer negligible improvements while increasing computational latency, suggesting that 8 coupling layers provide sufficient transformation capacity for modeling pose residuals.

\section{Conclusion and Future Work}

In this paper, we introduce MSG-Flow, a hierarchical framework for skeleton-based VAD that addresses key challenges in embodied perception. By explicitly modeling motion at discrete semantic and continuous kinematic levels, MSG-Flow provides a principled approach to privacy-preserving anomaly detection in physical environments. The framework decomposes skeleton sequences into learned motion primitives through vector quantization, captures temporal dependencies among primitives using autoregressive Transformers, and simultaneously models fine-grained pose variations through conditional normalizing flows. 
Extensive experiments on benchmarks demonstrate SOTA performance, 
and comprehensive ablation studies confirm the complementary contributions of semantic and kinematic modeling components. The framework maintains computational efficiency suitable for resource-constrained embodied systems. 

Future work will explore several promising directions to advance privacy-preserving VAD in embodied multimedia systems. To be specific, extending the framework to model inter-person interactions would enable detection of group-level anomalies in collaborative or adversarial scenarios. Moreover, investigating privacy-aware appearance integration could selectively incorporate visual cues while maintaining privacy guarantees, potentially improving detection in appearance-sensitive anomaly types. 

\bibliographystyle{IEEEtran}
\bibliography{refs.bib}

\end{document}